\documentclass{article} 
\usepackage[table,xcdraw]{xcolor}

\usepackage{lmrl2025_workshop,times}

\usepackage{amsmath,amsfonts,bm}

\def\eqref#1{equation~\ref{#1}}

\def\1{\bm{1}}

\DeclareMathAlphabet{\mathsfit}{\encodingdefault}{\sfdefault}{m}{sl}
\SetMathAlphabet{\mathsfit}{bold}{\encodingdefault}{\sfdefault}{bx}{n}

\usepackage{hyperref}
\usepackage{url}
\usepackage{booktabs}
\usepackage{multirow}
\usepackage{float}
\usepackage{xfrac}

\usepackage{graphicx}
\usepackage{subcaption}
\usepackage{caption}
\usepackage{amsmath,amsfonts,amssymb}

\def\cN{\mathcal{N}}

\author{%
  Eoin Quinn\thanks{e.quinn@instadeep.com, o.bent@instadeep.com} , Ghassene Jebali, Maxime Seince, Oliver Bent$^*$\\
  InstaDeep Ltd, \textit{42 rue de Paradis, 75010 Paris, France}
}

\title{Discriminative protein sequence modelling with Latent Space Diffusion} 

\lmrlfinalcopy 
\begin{document}

\maketitle

\begin{abstract}


We  explore a framework for protein sequence representation learning that decomposes the task between manifold learning and distributional modelling.
Specifically we present a Latent Space Diffusion architecture which combines a protein sequence autoencoder with a denoising diffusion model operating on its latent space. We obtain a one-parameter family of learned representations from the diffusion model, along with the autoencoder’s latent representation. 
We propose and evaluate two autoencoder architectures: a homogeneous model forcing amino acids of the same type to be identically distributed in the latent space, and an inhomogeneous model employing a noise-based variant of masking.
As a baseline we take a latent space learned by masked language modelling, and evaluate discriminative capability on a range of protein property prediction tasks. Our finding is twofold: the diffusion models trained on both our proposed variants display higher discriminative power than the one trained on the masked language model baseline, none of the diffusion representations achieve the performance of the masked language model embeddings themselves.

\end{abstract}

\section{Introduction}

Proteins are an important class of biomolecules whose function, interaction, and evolutionary relationships are central to understanding cellular mechanisms and the complexity of life. 
While the underlying principles governing proteins and their behaviour admit explicit formulations in quantum chemistry, these are in practice too complex to model directly. The quest for simplifying representations and approximations which strike a balance between generality, accuracy and computational efficiency is a core challenge of computational biology.


Machine learning provides a powerful suite of tools for representation learning. A prominent method is an autoencoder employing an information bottleneck to learn a compressed latent representation \cite{tishby2000information}.  
For proteins this is less applicable however, as a protein's primary sequence already provides an incredibly compact representation (each amino acid, being a categorical variable of size 20, can be represented with just 5 bits). Indeed the primary sequence completely determines a protein, and the key challenge is how to decode from this. 
Applications of machine learning to protein representation learning from sequence data can be roughly organised around two main threads: those such as AlphaFold which leverage multiple sequence alignments (MSAs) capturing co-evolutionary information \cite{alphafold, rao2021msa, truong2023poet}, and those such as ESM \cite{esm, esm2, esm3} utilising masked language modelling (MLM) \cite{devlin-etal-2019-bert, elnaggar2021prottrans, brandes2022proteinbert}.

Generative modelling is closely tied to representation learning \cite{Kingma2014, goodfellow2014generative}. Indeed masked language modelling is a form of reconstructive learning, where a model is trained to restore partially corrupted input, which underlies its ability to learn rich contextually aware representations \cite{devlin-etal-2019-bert}. 
For continuous spaces,  Gaussian diffusion has emerged as a leading generative method due to its ability to produce diverse high-quality samples from complex distributions \cite{sohl2015deep, ho2020denoising, song2020score}. By learning how Gaussian noise diffuses through a data space, a diffusion model learns to approximate the score function of the data distribution, $s(x) = \nabla_x\log p_{data}(x)$. From a statistical physics perspective, expressing the distribution in Boltzmann form  $p_{data}(x)\sim \exp(-E(x))$, the score function admits a natural interpretation as a distributional force: $F(x) \equiv -\nabla_x E(x) = s(x)$, which underlies a diffusion model's ability to both navigate the distribution efficiently as well as to learn a meaningful representation of the data \cite{song2020score, vincent2008extracting}.

Studies of generative modelling on protein sequence data have primarily focused on discrete diffusion methods \cite{gruver2024protein, evodiff, DPLM}. These are counterparts to Gaussian diffusion that place emphasis on the categorical nature of amino acids. 
In contrast to the continuous setting however, the formalism of discrete diffusion is less well established, and  principled approaches have appeared only relatively recently \cite{campbell2024generative, shi2024simplified, zheng2024masked, gat2024discrete}. Another notable work is DPLM \cite{DPLM}, an adaption of masked language modelling, where in particular the authors highlight the representation learning capabilities of their generative model and demonstrate it achieves competitive performance with ESM2.

Let us question however  whether a masking-based approach is the best route towards modelling protein sequence data. From a reconstructive learning perspective, it is unclear whether masking is an optimal way to represent a corrupted sequence. For instance, the unmasked amino acids are fully specified with no ambiguity, while for the masked amino acids only their ambiguity is conveyed. One can imagine instead an alternative corruption process where partial information is retained/erased, for example expected physiochemical properties or aspects of long-range dependencies. Similarly from a generative perspective, one may question the task of performing distributional modelling directly in the discrete domain, as at the level of sequence the protein landscape is far from smooth: single mutations can have abrupt consequences while compound mutations may be strongly correlated.

This motivates us to explore a switch in focus from a discrete representation of sequence space to a continuous one. This can be framed as making a distinction between two aspects of protein representation learning: manifold learning and distribution learning. The first addresses the question of how to embed protein sequences in a continuous latent space, while the second concerns the distribution of protein sequences over this latent space. Here Gaussian diffusion can be employed for the distributional modelling, 
and so the question then is how to learn an appropriate latent space.

Previous works adopting a latent diffusion approach for protein sequence data examined the use of the ESM embeddings \cite{ESM-diffusion, AMP-diffusion, lu2024generating}. There is a difficulty with this approach however,
which can be attributed to the embeddings retaining much of the discreteness of the underlying sequence \cite{li2023diffusion}. In essence, amino acid embeddings are too robust to added noise, which obstructs the learning ability of denoising. Indeed a parallel can be made here to high-resolution images, for which latent diffusion models were originally introduced \cite{rombach2022high}.

In this work we attempt to address the challenge of how to construct a latent space which facilitates the distributional modelling of proteins sequence data. To this end we propose two novel sequence autoencoder architectures: a homogeneous model forcing amino
acids of the same type to be identically distributed in the latent space, and an inhomogeneous model employing a noise-based variant of masking. We train a diffusion model on their latent space, and identify how this gives rise to an additional one-parameter family of learned representations. We focus on this discriminative capability of the diffusion model, and evaluate it on a diverse set of representatation learning benchmarks.

\section{Related work}

The use of diffusion/denoising for protein representation learning was introduced for the structural representation \cite{zaidi2022pre, liu2022molecular}, based on a connection between the learned score function and molecular force fields. DSMBind employs SE(3) denoising score matching as an unsupervised pre-training task for binding energy prediction \cite{DSMBind}. From a generative perspective, diffusion has also been mostly applied to structure~\cite{RFdiffusion, chroma, framediff, lee2023score}, for which the Gaussian form of diffusion can be employed. On protein structure prediction,  AlphaFold 3 \cite{AF3} trains a conditioned diffusion model for the generation of its structural predictions,  and PLAID \cite{lu2024generating}  takes a latent diffusion approach. 
LatentDock \cite{mcpartlon2023latentdock} employs structure-based latent diffusion for modelling protein-protein docking.
Application to the generation of conformational ensembles has also been explored \cite{jing2024alphafold, hassan2024flow}.

Discrete diffusion applied on protein sequence data has been explored in LaMBO-2 \cite{gruver2024protein}, EvoDiff \cite{evodiff}, and DPLM \cite{DPLM}. DPLM stands out for evaluating the representation learning capabilities of their model, demonstrating it's ability to perform competitively across a range of prediction tasks. There are also works exploring latent diffusion on language model embeddings. DiMA \cite{ESM-diffusion} and AMP-Diffusion (focused solely on antimicrobial peptides) \cite{AMP-diffusion} employed pre-trained ESM2 embeddings, and PLAID \cite{lu2024generating} employed the compressed CHEAP \cite{CHEAP} embeddings of ESM2. Both DiMA and PLAID trained their diffusion models with the  heuristic method of self-conditioning \cite{self-conditioning}.

Two studies explored latent diffusion on pre-trained ESM2 embeddings \cite{AMP-diffusion, ESM-diffusion}, and see also PLAID \cite{lu2024generating} which employs the compressed CHEAP \cite{CHEAP} embeddings of ESM.

More broadly, the link between discriminative and generative modelling  underlies the auto-regressive approach \cite{protgpt, progen}, the masked language modelling approach \cite{esm, esm2, esm3,  elnaggar2021prottrans, brandes2022proteinbert}, as well as variational autoencoder approaches \cite{sinai2017variational, sevgen2023prot}, to modelling protein sequence data.

Beyond the field of protein modelling, denoising autoencoders date back to the seminal work \cite{vincent2008extracting}. 
Diffusion-based representation learning was advanced in \cite{abstreiter2021diffusion}. 
Latent diffusion models were introduced in the image domain \cite{rombach2022high}, and the application of latent diffusion to discrete data has been predominantly studied in the natural language processing literature \cite{Diffusion-lm, dieleman2022continuous, strudel2022self, gao-etal-2024-empowering, DINOISER, gulrajani2024likelihood}.

\section{Latent Space Diffusion}\label{sec:architecture}

\begin{figure}
  \centering
  \includegraphics[width=0.85\textwidth]{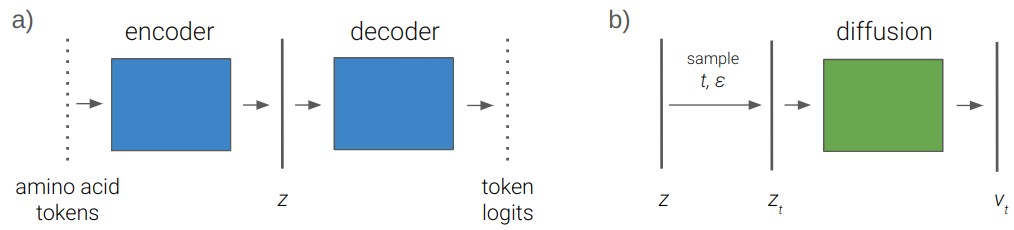}
  \caption{The LSD model is comprised of \textbf{(a)} a protein sequence autoencoder which learns a latent space $z$, and \textbf{(b)} a diffusion model acting on this latent space. 
  The autoencoder is trained end-to-end by balancing a reconstruction loss, between input amino acid tokens and output token logits, against a normalization loss
on the distribution of  latent space embeddings. We consider two variants,  LSD-TN with a non-trivial normalization loss and LSD-NM with a non-trivial reconstruction loss, as described in Sec.~\ref{sec:architecture}.
  The diffusion model learns to map noised latent embeddings $z_t = \cos(\pi t/2) z + \sin(\pi t/2) \varepsilon$ to their orthogonal complements $v_t = -\sin(\pi t/2) z + \cos(\pi t/2) \varepsilon$, which thereby provides an additional one-parameter family of sequence representation to that obtained at the latent space.}
  \label{fig:architecture}
\end{figure}

We employ a Latent Space Diffusion (LSD) architecture, illustrated in Fig.~\ref{fig:architecture}, comprised of 
\begin{itemize}
    \item an autoencoder learning a latent manifold embedding of protein sequences,
    \item a diffusion model for the distributional modelling of protein sequence datasets over this learned latent space. 
\end{itemize}
We adopt a transformer-based architecture for each component as shown in Appendix \ref{app:model}.
The autoencoder is composed of an encoder-decoder pair, which we set to have an equal number of layers. The encoder takes as input tokenized protein sequences, and outputs latent embeddings $z_{a,i}$ of amino acids, with $a$ indexing amino acid position and $i$ the coordinate of the embedding space. The decoder takes the latent embeddings $z_{a,i}$ as input, and outputs corresponding token logits. The diffusion model is a conditioned transformer and we describe its action on the latent space below in the description of the diffusion loss. A more detailed description of the architecture is provided in Appendix~\ref{app:model}.

We train the auto-encoder under competition between a reconstruction loss and a normalization loss on the latent embeddings. For the reconstruction loss we employ the standard cross-entropy  between the input tokens and output token logits. The normalization loss is less straightforward.
In contrast to a variational auto-encoder \cite{Kingma2014} which maps input to distributions over the latent space for which it learns the mean and variance, we let the encoder map directly to the latent space and guide the distribution of $z_{a,i}$ over $a$ to be normally distributed (this foregoes the generative capability of the autoencoder, which is compensated for here by the diffusion model). 
Specifically, given a batch of sequences we employ a univariate parametric form of the Kullback–Leibler divergence 
\begin{equation}
\mathcal{L}_N = \frac{1}{2d}\sum_{i=1}^d (\mu_i^2 + \sigma_i^2 - \log \sigma_i^2 -1 ),
\end{equation}
expressed in terms of the empirical mean $\mu_i$ and variance $\sigma_i^2$ of the $z_{a,i}$, with $d$ the embedding dimension. We considered also a multivariate parametric form but found this to degrade performance, see ablation in Appendix~\ref{app:ablation}.

The simple combination of reconstruction loss and normalisation loss is not however sufficient to drive meaningful learning in the latent space.
To achieve this, we consider two variants as follows:
\begin{itemize}
    \item \textbf{Token Norm (LSD-TN):} 
    here we modify the normalization loss by applying it separately to the embeddings of each amino acid type. Specifically, within each batch, we partition the latent embeddings into 20 sets, each corresponding to one of the 20 canonical amino acids, and employ a separate normalization loss for each set. While the default normalization loss allows different amino acid types to occupy distinct regions under the same normal distribution, this approach imposes a stricter constraint, creating an effective bottleneck for representation learning.
    
    \item \textbf{Noise Masking (LSD-NM):} here we modify the reconstruction loss to a variant of MLM designed for greater robustness to noise. Unlike standard MLM, where a fraction of amino acid embeddings are fully masked while the rest remain unaltered, our approach applies varying levels of corruption by inhomogeneously adding Gaussian noise to the latent embeddings. Specifically, we transform each amino acid embedding vector as
    \begin{equation}
    z_a \to  \cos(\pi t_a/2) z_{a} + \sin(\pi t_a/2) \varepsilon,
    \end{equation}
    where $t_a\in(0,1)$ controls the noise level and $\varepsilon$ is an embedding vector sampled from $\mathcal{N}(0,1)$. To reflect this corruption in the reconstruction loss, we weight each embedding’s contribution by the noise amplitude $\sin^2(\pi t_a/2)$, ensuring that highly corrupted embeddings dominate the training signal, while minimally corrupted ones contribute negligibly.

    We explored two sampling strategies for $t_a$: uniform sampling over (0,1) and sampling proportional to the signal amplitude  $\cos^2(\pi t_a/2)$, as used for training  the diffusion model (see below). The latter approach, which results in most amino acids being weakly noised while a few are strongly noised, performed better, and we adopt this choice in the models we present. See ablation in Appendix~\ref{app:ablation}.

    We emphasize that this inhomogeneous noising of the latent vectors is applied only during the training of the autoencoder. At inference the encoder maps deterministically to the latent space.
\end{itemize}

The diffusion model is trained on the autoencoder's latent space. 
For this we employ a variance-preserving cosine noise schedule \cite{nichol2021improved}, 
the $v$-target objective \cite{v-target}, and epsilon prediction loss \cite{ho2020denoising}, adopting the standard conventions of e.g. \cite{hoogeboom2023simple}. 
Specifically, the latent embeddings $z$ get (here uniformly) noised to 
\begin{equation}
z_t = \cos(\pi t/2) z + \sin(\pi t/2) \varepsilon,
\end{equation}
for $t\in(0,1)$  and $\varepsilon\in\mathcal{N}(0,1)$, and the diffusion model is trained to learn 
\begin{equation}
v_t = -\sin(\pi t/2) z + \cos(\pi t/2) \varepsilon.
\end{equation}
i.e.~$\hat{v}_t = \mathtt{Diffusion}(z_t, t)$.
The epsilon prediction loss, expressed in terms of $v$, is weighted by the signal amplitude 
\begin{equation}
\mathcal{L}_D =
\frac{1}{2} \mathbb{E}_{t\sim(0,1),\,\varepsilon\sim \cN(0,1)} \cos^2(\pi t/2) \| \hat{v}_t - v_t\|^2,
\end{equation}
and we evaluate this with importance sampling. 
Indeed, from a representation learning perspective the increased weight for sampling t closer to 0 is intuitive, as information gets washed out with increasing $t$.

In this work we focus on the discriminative capability of the diffusion model. While an ultimate objective of the LSD construction is to develop also the generative capability, we take the perspective that the discriminative capability serves as a useful guide for identifying an appropriate autoencoder architecture, and so defer the more challenging generative aspect until this is established. This choice is grounded on the intuition that a generative model can meaningfully generate only to the extent that it can discriminate.


Through its $t$-dependence, the diffusion model provides a one-parameter family of learned representations. There are two subtleties to this however. The first arises from the fact that the input to the diffusion model, $z_t$, depends on both $t$ and sampled $\varepsilon$. As $\varepsilon$ essentially amounts to Gaussian broadening, we can treat this as regularization and employ the mean value, i.e.~take 
\begin{equation}
\bar{v}_t(z) = \mathtt{Diffusion}(\cos(\pi t/2) z, t).
\end{equation}
The second subtlety is that the diffusion model learns nothing for $t\sim1$ as the input there is noise. This can be compensated for by switching to the score function, which from Tweedie's formula \cite{tweedies} is expressed as
\begin{equation}
    s_t(z) = - \frac{\hat{\varepsilon}_t(z)}{\sin(\pi t/2)} = -\frac{2}{\sin(\pi t)} \big(\hat{v}_t(z) + \sin(\pi t/2) z\big).
\end{equation}
Dropping the singular prefactor, we thus take the diffusion representations as 
\begin{equation}
\bar{v}_t(z) + \sin(\pi t/2) z. 
\end{equation}
At $t=0$, this reduces back to $\hat{v}_t(z)$.

\section{Evaluation}
\label{sec:evaluation}

\begin{table}[t]
    \centering
    \begin{tabular}{l|c|c}
        \toprule
        Model & Encoder / Decoder & Diffusion \\
        \midrule
        S      & 4.7M  & 7.3M  \\
        M      & 18.9M  & 29.0M  \\
        \bottomrule
    \end{tabular}
    \caption{Number of parameters for the S and M versions of the LSD model. The decoder is trivialised for the MLM diffusion baseline.}
    \label{table:model_size}
\end{table}

We present here two trained models for both LSD-TN and LSD-NM variants. We call these S and M, and provide their parameter counts in Table~\ref{table:model_size}.
The models are trained on the Uniref50 protein sequence dataset \cite{uniref50}, with sequences of maximum length 254, and omitting sequences with unknown or non-canonical amino acids ($0.5\%$ of the dataset).
Full model hyperparameters and further training details are given in  Appendix~\ref{app:model}.

To establish a baseline for their performance we additionally train corresponding MLM models, along with a diffusion model on their learned embeddings, using an identical setup. 
These MLM baseline models can be viewed as counterparts to DiMA \cite{ESM-diffusion}, with self-conditioning deactivated.
In terms of Fig.~\ref{fig:architecture}, for these MLM models the decoder's transformer trunk is trivialized. Masking is applied at the input to the encoder, as opposed the input of the decoder for LSD-NM. We employ a 15\% masking rate, and extract the latent embeddings after the layer norm following the transformer layers to ensure they are appropriately normalised.


We assess the discriminative capabilities of these models across a set of a property prediction tasks assessing stability, interaction and functional characterization, which we adopt from SaProt \cite{su2023saprot}. 
For this we freeze the backbone and training a simple predictor on the mean of the embeddings across the sequence. We provide further information on the datasets and predictor architecture in Appendix \ref{app:evaluation_tasks}.

We report the performance of the models in Table \ref{table:eval}. For the LSD-NM and LSD-TN models and the MLM diffusion baseline, we evaluate on both the latent space, at the output of the encoder, and on the $t=0$ output of the diffusion model applied on the latent space. To further benchmark these results we additionally evaluate the predictor on two prominent protein representation learning models, ESM2 \cite{esm2} and the discrete diffusion model DPLM \cite{DPLM}.

\begin{table}[t]
    \centering
   \resizebox{\textwidth}{!}{
    \begin{tabular}{@{}lccccc@{}}
        \toprule
        \multirow{3}{*}{Models} & \multirow{2}{*}{Thermostability $\uparrow$} & \multirow{2}{*}{HumanPPI $\uparrow$} & \multirow{2}{*}{Metal Ion Binding $\uparrow$} & \multicolumn{2}{c}{DeepLoc $\uparrow$} \\
        \cmidrule(lr){5-6}
        & &  &  & Subcellular & Binary  \\
        \cmidrule(lr){2-6}
        & Spearman's $\rho$ & Acc (\%) & Acc (\%) & Acc (\%) & Acc (\%)  \\
        \midrule
        ESM 8M    & 0.648 & 72.7 & 63.2 & 68.2 & 88.8 \\
        ESM 650M  & 0.690 & 81.3 & 66.8 & 77.6 & 91.0 \\
        DPLM 650M & 0.693 & 76.7 & 69.1 & 78.5 & 90.8 \\ 
        \midrule
        MLM-S: Encoder                                    & 0.606 & 72.3 & 65.0 & 60.1 & 86.3 \\
        \rowcolor[HTML]{e3e8ee} MLM-S: Diffusion ($t=0$) & 0.474 & 57.7 & 63.0 & 46.1 & 74.3 \\ 
        MLM-M: Encoder                                   & 0.613 & 72.3 & 63.5 & 62.4 & 87.3 \\
        \rowcolor[HTML]{e3e8ee} MLM-M: Diffusion ($t=0$) & 0.543 & 60.6 & 61.7 & 52.2 & 76.2 \\ \midrule
                                LSD-TN-S: Encoder   & 0.560 & 58.6 & 64.6 & 53.0 & 76.6 \\
        \rowcolor[HTML]{d5f9e8} LSD-TN-S: Diffusion ($t=0$) & 0.562 & 62.6 & 62.8 & 48.2 & 75.3 \\
                                LSD-TN-M: Encoder  & 0.571 & 59.1 & 63.2 & 54.4 & 76.7 \\
        \rowcolor[HTML]{d5f9e8} LSD-TN-M: Diffusion ($t=0$) & 0.571 & 65.9 & 62.6 & 52.7 & 76.5 \\ \midrule
                                LSD-NM-S: Encoder   & 0.553 & 62.6 & 64.1 & 54.6 & 77.6 \\
        \rowcolor[HTML]{d5f9e8} LSD-NM-S: Diffusion ($t=0$) & 0.567 & 60.2 & 65.0 & 53.5 & 76.1 \\
                                LSD-NM-M: Encoder  & 0.571 & 61.6 & 64.6 & 55.0 & 77.3 \\
        \rowcolor[HTML]{d5f9e8} LSD-NM-M: Diffusion ($t=0$) & 0.581 & 61.1 & 64.7 & 54.2 & 76.8 \\
        \bottomrule
    \end{tabular}%
   }
    \caption{Evaluation on protein property prediction tasks. 
    The $t=0$ diffusion representations are highlighted, green for the LSD models and gray for the MLM baseline. Reported scores are computed as the  mean of 5 randomly initialized predictors. 
    }
    \label{table:eval}
\end{table}

We first highlight the diffusion model results, which are the primary focus of this work. We see that the LSD-TN and LSD-NM diffusion models consistently outperform the MLM diffusion models across all evaluation metrics. 
Comparing the between the LSD-TN and LSD-NM variants, we see that the LSD-NM performs better than  LSD-TN on all but one task. The exception is the HumanPPI, on which LSD-TN-M performs notably better. This may indicate a complementarity in how the two different constructions organise correlations within their respective latent spaces. 

Turning to the latent representations of the encoder we see that the situation is reversed, with the MLM results here greatly outperforming their LSD counterparts. This aligns with the well-established strength of masked language modelling for representation learning, in contrast to the LSD autoencoders which were not designed to optimise for this. Indeed, the MLM encoder performs best of all the evaluated MLM, LSD-TN and LSD-NM representations, and performs significantly better than even the best LSD diffusion models. This trend is also exhibited by the ESM 8M model, which has a smaller parameter count than all the LSD models.

We also compare between the encoder and diffusion representations. For the LSD-TN model we observe a complementarity, with results consistently better for the encoder on Metal Ion Binding and DeepLoc-Subcellular, on a par for Thermostability and DeepLoc-Binary, and better for the diffusion model on HumanPPI. For LSD-NM the results are more similar between the two modules, while for the MLM model the diffusion representations all significantly score lower than the encoder's.

\begin{figure}[t]
    \centering
    \includegraphics[width=0.85\linewidth]{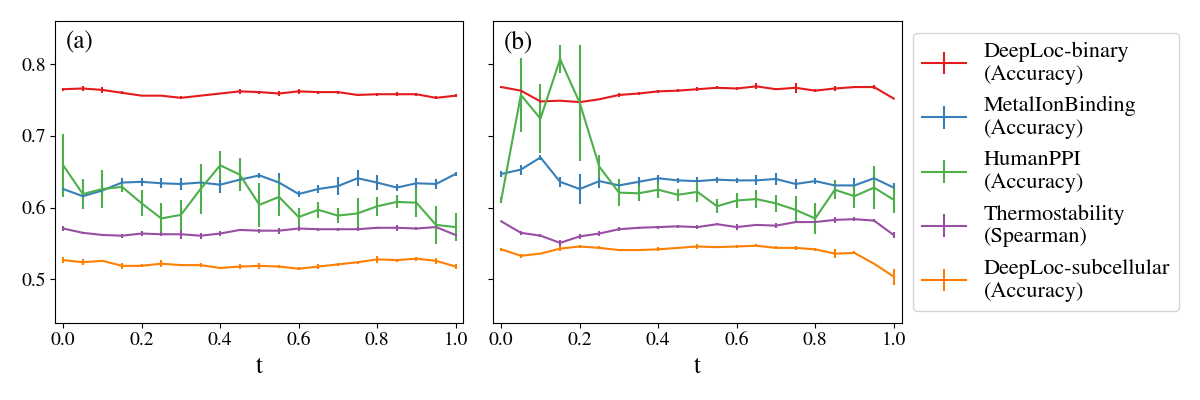}
    \caption{
    Evaluation of the $t$-dependence of the diffusion representation for the five protein property prediction tasks: \textbf{(a)} LSD-TN-M, \textbf{(b)} LSD-NM-M. The error bars are computed from the results of 5  randomly initialized predictors.}
    \label{fig:diffusion_t}
\end{figure}

We now turn to the $t$-dependence of the diffusion representations. In Fig.~\ref{fig:diffusion_t} we evaluate the performance of the regularised score function $\bar{v}_t(z) + \sin(\pi t/2) z$ for all five tasks for the LSD-TN-M and LSD-NM-M models. For the LSD-TN model we observe that the curves are notably flat, with the exception of HumanPPI although that could reflect the greater uncertainty in that metric. For LSD-NM on the other hand, there is some variation to the curves with different trends for the different tasks. This may reflect the expectation that different correlations are captured at the different scales parameterised by $t$, but a definitive conclusion cannot be made. Again the HumanPPI metric stands out. We observe a consistent peak at $t=0.15$ with value $0.807\pm0.019$, which (remarkably) is on a par with the ESM2 and DPLM 650M models. Given the inconsistent behaviour of the HumanPPI evaluation across all experiments however, it is difficult to conclude how meaningful this result is.

\subsection{Visualization}

\begin{figure}[t]
    \centering
    \includegraphics[width=0.99\linewidth]{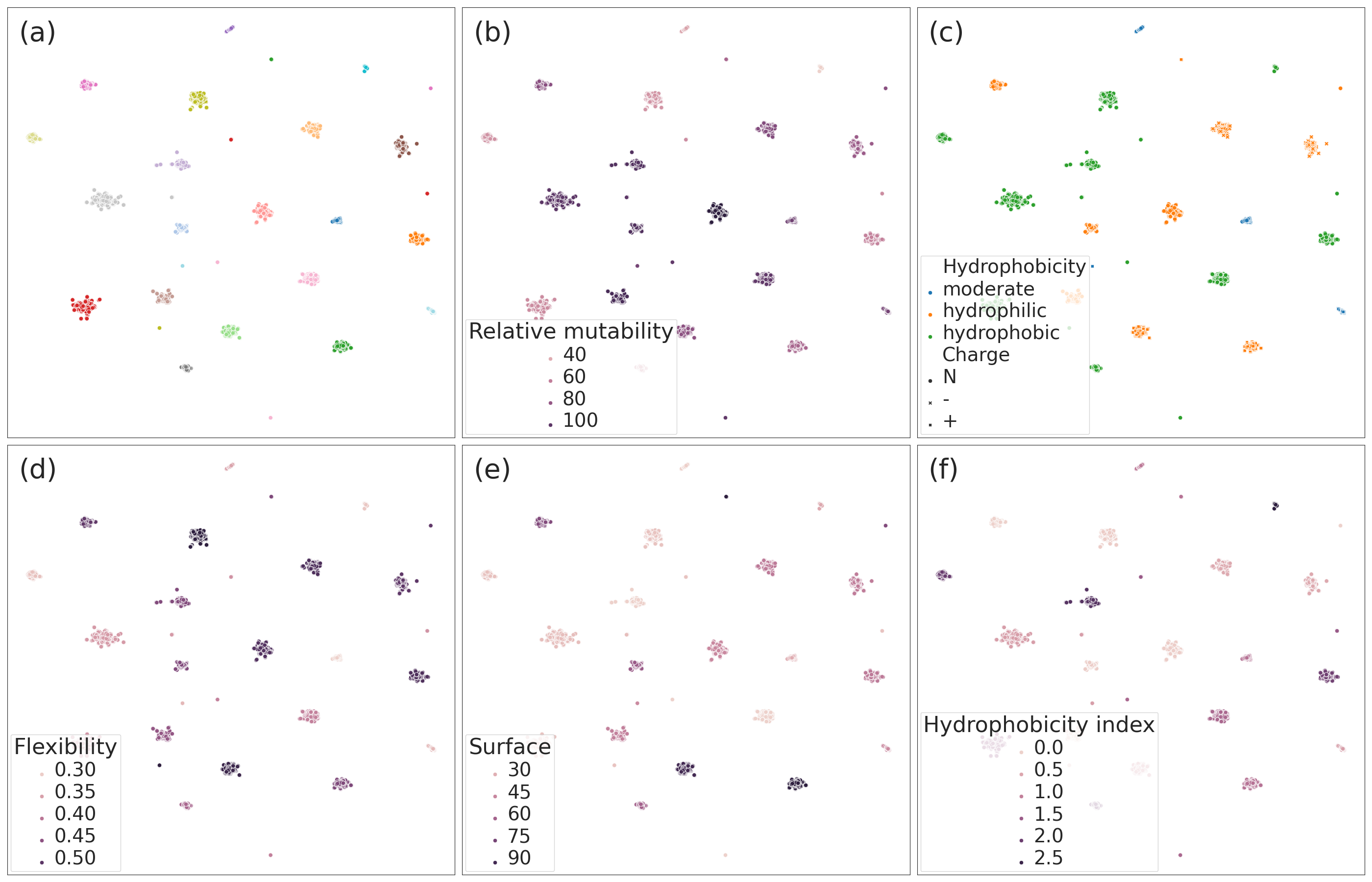}
    \caption{UMAP projections of the latent space learned by LSD-NM-M Diffusion model. \textbf{(a)} Coloured by amino acid. \textbf{(b)} Coloured by relative mutability \cite{jones1992rapid}. \textbf{(c)} Coloured by hydrophobicity nature. \textbf{(d)} Coloured by hydrophobicity index \cite{argos1982structural}. \textbf{(e)} Coloured by average flexibility index \cite{bhaskaran1988positional}. \textbf{(f)} Coloured by residue accessible surface area in folded protein \cite{chothia1976nature}.
    }
    \label{fig:umap}
\end{figure}

To complement the above quantitative analysis, we provide a UMAP-based visual analysis of the learned representations in Fig.~\ref{fig:umap}. We focus on the best performing LSD-NM diffusion representation, and use colouring to highlight the learned biological features. For each plot, we sample 64 sequences of length 100 amino acids from UniRef50, process them through the encoder and diffusion models, and employ UMAP to project the resulting embeddings to 2D.

We also conduct an attention map analysis to visualise how contextual information is integrated, and present this in appendix \ref{app:attention_map}.

\section{Discussion}\label{sec:discussion}

The results of our evaluation highlight the key challenge in applying latent space diffusion to protein sequences: identifying an appropriate latent space. We observe that embeddings optimized for representation learning, e.g. those from the MLM baseline, result in an underperforming diffusion model. To address this, we proposed and analyzed alternative latent space learning methods designed to prioritise well-distributed embeddings. While these achieved the goal of boosting the diffusion model's performance, they ultimately fell short of matching the overall performance of token-based reconstructive learning methods like MLM, or the discrete diffusion method of DPLM.

Nevertheless, the autoencoder architectures we present have interesting features that may warrant further study. To our knowledge, the Token Norm bottleneck introduced here is novel. It is particularly suitable for protein sequence data, where the 20 amino acids provide a limited vocabulary compared to the much larger token vocabularies commonly used in NLP sequence modelling. The LSD-TN model is notable for its simplicity, achieving reasonable representation  learning performance despite possessing a homogeneous bottleneck. We remark also that the univariate parametric form of the Kullback-Leibler divergence normalization loss is crude, and can perhaps be improved.

Our noise masking strategy for the LSD-NM model is quite similar to the diffusion denoising. A key difference however is that for the autoencoder the noise is applied  inhomogeneously, while for the diffusion model it is applied uniformly across the sequence. The former places more emphasis on locality, while the latter learns the overall data distribution, underpining its generative capability. It may be interesting to explore if the two can be effectively combined. One possibility  is to let the decoder and diffusion model share the same transformer trunk. 

We comment also on the one-parameter representations offered by the $t$-dependence of the diffusion model. As described in Sec.~\ref{sec:architecture} these amount to a regularised form of the score function, $s_t(z)$. We recall from the Introduction that the score function admits an interpretation as a distributional force. Given that our models are trained on the evolutionary-scale Uniref50 dataset, we can thus offer an interpretation of $s_t(z)$ as a representation of the forces governing proteins, with the parameter $t$ setting the scale of the latent space over which these forces are computed. 

The DPLM representation included in Table~\ref{table:eval} correspond to the $t=0$ pass of their discrete diffusion model. It is unclear whether these can be extended non-zero $t$, as in this case the self-averaging property of Gaussian noise is lost, but this may be worthy of further investigation.

\section{Conclusion and Outlook}

We have presented a Latent Space Diffusion approach for modelling protein sequence data, with an initial focus on discriminative modelling. 
We highlighted the key challenge in developing this framework, which is to learn a sufficiently well-distributed latent space for the effective training of the diffusion model. 
To this end we proposed two novel autoencoder architectures: LSD-TN and LSD-NM. 
We evaluated their  predictive performance across a range of protein prediction tasks and conducted an ablation study of key design choices. 
We found that while the diffusion performed better than with an MLM baseline, ultimately our trained models underperformed relative to token-based reconstructive learning approaches.

Our study provides an initial exploration of the LSD approach and opens up several interesting directions for future work. The architecture itself is not settled, and further research is needed to refine the question of what constitutes a good latent space. It remains unclear whether the inherent discreteness of sequence data can be sufficiently removed from the latent space to justify reconstructive learning through denoising.  Another aspect is the latent space's dimensionality. It has been demonstrated that the learned embeddings of ESM2 can be massively compressed without significantly degrading their information content \cite{CHEAP}. This motivates a compression of the latent space, which in turn can facilitate more effective distributional modelling.

There is much to be learned about the richness of the information captured by the one-parameter family of diffusion representations, and how this can be best employed for protein modelling. Looking ahead, we also highlight that LSD could serve as a pre-trained model for fine-tuning on specific tasks. In particular, freezing the autoencoder while fine-tuning the diffusion model offers a particularly natural route forward, and may help to bypass the catastrophic forgetting issues observed in masked language models \cite{wallat2021bertnesia,schmirler2024fine}.

Another promising avenue is the incorporation of additional modalities, particularly protein structural data \cite{mansoor2024protein}. In this regard the continuous nature of the LSD formulation provides an advantage over discrete token-based approaches \cite{esm3, su2023saprot}.

Finally, it would be of great interest to scale up the model and explore its generative capabilities.






\bibliographystyle{iclr2025_conference}

\appendix

\section{Additional model details}\label{app:model}

\begin{figure}[H]
    \centering
    \begin{subfigure}{0.32\textwidth}
        \centering
        \includegraphics[width=\linewidth]{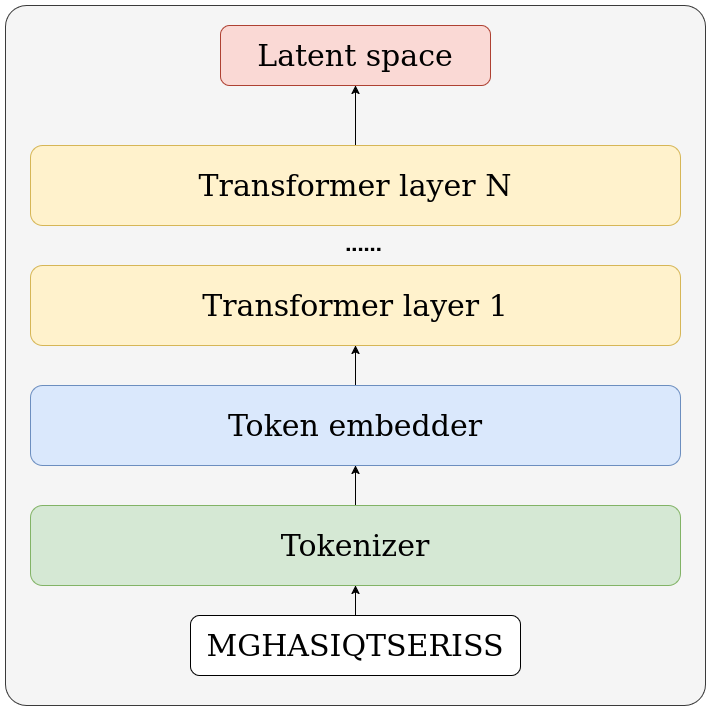}
        \caption{}
        \label{fig:encoder}
    \end{subfigure}
    \hfill
    \begin{subfigure}{0.32\textwidth}
        \centering
        \includegraphics[width=\linewidth]{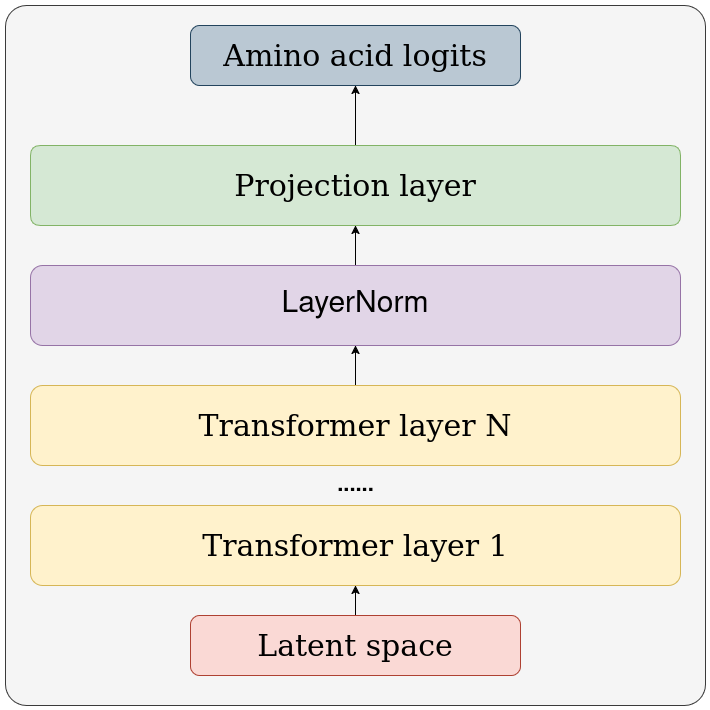}
        \caption{}
        \label{fig:decoder}
    \end{subfigure}
    \hfill
    \begin{subfigure}{0.32\textwidth}
        \centering
        \includegraphics[width=\linewidth]{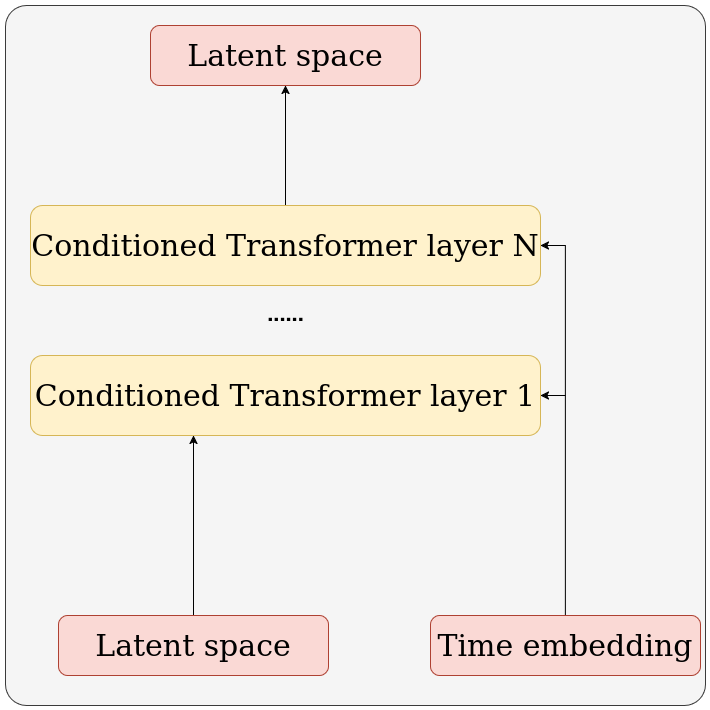}
        \caption{}
        \label{fig:diffusion}
    \end{subfigure}
    
    \caption{Detailed architecture: \textbf{(a)} encoder, \textbf{(b)} decoder \textbf{(c)} diffusion model.}
    \label{fig:model_architecture}
\end{figure}

All three components of the model, encoder-decoder-diffusion, are based on the transformer architecture as illustrated in Fig.~\ref{fig:model_architecture}.
We employ RoPE positional encoding \cite{RoPE}, Pre-LN \cite{preLN}, SwiGLU activation functions \cite{swish}. Sequences are padded to maximum sequence length 256, and each component has internal BOS/EOS embeddings which are learned independently and not output. 
For the decoder we employ a simple projection head onto the amino acid logits.
Time-conditioning for the diffusion model is implemented using the adaLN-zero prescription described in DiT~\cite{peebles_scalable_2023}.

The models are trained on Uniref50 with sequences limited to a maximum length of 254 (+2 for BOS/EOS embeddings), with a batch size of 512, using the AdamW optimizer with weight decay 1e-3 and learning rate 2e-5, on one A100 80GB GPU.
Model hyper-parameters are provided in Table~\ref{table:model_details}.

\begin{table}[H]
    \centering
    \caption{Model details}
    \label{table:model_details}
    \begin{tabular}{llccccc}
        \toprule
        Model name & Module & Model size & Channels & Heads & Layers & Steps\\
        \midrule
        \multirow{2}{*}{MLM-S}  & Encoder & 4.7M  & \multirow{2}{*}{256} & \multirow{2}{*}{16} & \multirow{2}{*}{6} & \multirow{2}{*}{200k}\\
                                & Diffusion       & 7.3M  &  &  &  & \\
        \midrule
        \multirow{2}{*}{MLM-M}  & Encoder & 18.9M & \multirow{2}{*}{512} & \multirow{2}{*}{16} & \multirow{2}{*}{6} & \multirow{2}{*}{100k}\\
                                & Diffusion       & 29.0M &  &  &  & \\
        \midrule
        \multirow{2}{*}{LSD-TN-S} & Encoder/Decoder & 4.7M  & \multirow{2}{*}{256} & \multirow{2}{*}{16} & \multirow{2}{*}{6 } & \multirow{2}{*}{200k}\\
                                  & Diffusion       & 7.3M  &  &  & & \\
        \midrule
        \multirow{2}{*}{LSD-TN-M} & Encoder/Decoder & 18.9M & \multirow{2}{*}{512} & \multirow{2}{*}{16} & \multirow{2}{*}{6} & \multirow{2}{*}{100k}\\
                                  & Diffusion       & 29.0M &  &  &  & \\
        \midrule
        \multirow{2}{*}{LSD-NM-S} & Encoder/Decoder & 4.7M  & \multirow{2}{*}{256} & \multirow{2}{*}{16} & \multirow{2}{*}{6} & \multirow{2}{*}{200k}\\
                                  & Diffusion       & 7.3M  &  &  &  & \\
        \midrule
        \multirow{2}{*}{LSD-NM-M} & Encoder/Decoder & 18.9M & \multirow{2}{*}{512} & \multirow{2}{*}{16} & \multirow{2}{*}{6} & \multirow{2}{*}{100k}\\
                                  & Diffusion       & 29.0M &  &  &  & \\
        \bottomrule
    \end{tabular}
\end{table}

\section{Evaluation details} \label{app:evaluation_tasks}

In Section~\ref{sec:evaluation} we evaluate representation learning on a set of protein property prediction tasks which we adopt from SaProt \cite{su2023saprot}:

\begin{itemize}

    \item \textbf{Thermostability}: 
    protein melting temperature $T_m$ data from the “Human-cell” splits of the Thermostability task of the FLIP  benchmark \cite{dallago2021flip}. 
    
    \item \textbf{HumanPPI}: 
    binary classification whether two proteins interact from HumanPPI data
    \cite{pan2010large} of the PEER benchmark \cite{xu2022peer}.
    
    \item \textbf{Metal Ion Binding}: binary classification of presence of metal ion–binding sites within a protein \cite{hu2022exploring}.

    \item \textbf{DeepLoc}: Predicts subcellular localization of proteins from the DeepLoc dataset \cite{almagro2017deeploc}. 
    \begin{itemize}
        \item \textbf{Subcellular}: 
        multi-class classification identifying one of 10 distinct subcellular compartments.
        \item \textbf{Binary}: binary classification between membrane-bound or soluble.
    \end{itemize}
\end{itemize}

We freeze the backbone and train an MLP predictor with a single hidden layer, and either a regressor or classifier head depending on the task,
The predictor acts on the mean of the embeddings across the sequence, and for the HumanPPI task we concatenate the mean embeddings of the two proteins. We set the dimension of the  predictor's hidden layer equal to that of the input.

This differs slightly from the SaProt pipeline (also employed by DPLM), which fine-tunes the backbone. As a result the values we obtain for ESM2 are not identical to theirs, nevertheless remain directly comparable.

\section{Ablations}\label{app:ablation}


\begin{table}[H]
    \centering
    \resizebox{\textwidth}{!}{
    \begin{tabular}{@{}lc|lccccc@{}}
        \toprule
        & &   & \multirow{2}{*}{Thermostability $\uparrow$} & \multirow{2}{*}{HumanPPI $\uparrow$} & \multirow{2}{*}{Metal Ion Binding $\uparrow$}& \multicolumn{2}{c}{DeepLoc $\uparrow$} \\
        \cmidrule(lr){7-8}
        Model & Importance sampling & Modules &  &  &  &  Subcellular & Binary  \\
        \cmidrule(lr){1-8}
        \multirow{6}{*}{LSD-NM-S}  & \multirow{2}{*}{$\checkmark$}  & Encoder   & 0.553 & 62.6 & 64.1 & 54.6 & 77.6 \\
                                   &                                & Diffusion & 0.567 & 60.2 & 65.0 & 53.5 & 76.1 \\
                                   \cmidrule(lr){2-8}
                                   & \multirow{2}{*}{Off for diffusion}  & Encoder   & 0.558 &  61.7& 63.7& 54.4 & 77.3 \\
                                   &                                    & Diffusion & 0.545 & 64.7& 63.8 &52.7 & 75.8 \\
                                   \cmidrule(lr){2-8}
                                   & \multirow{2}{*}{Off for noise masking}  & Encoder   & 0.543 & 60.6 &  62.8 &   55.2 &  77.2 \\
                                   &                                    & Diffusion &  0.540 & 68.0 &  59.2 &  52.2 & 76.5 \\
        \bottomrule
    \end{tabular}%
    }
    \caption{Importance sampling ablation.}
    \label{table:eval_abl_NM}
\end{table}


\begin{table}[H]
    \centering
    \resizebox{\textwidth}{!}{
    \begin{tabular}{@{}lc|lccccc@{}}
        \toprule
        & & & \multirow{2}{*}{Thermostability $\uparrow$} & \multirow{2}{*}{HumanPPI $\uparrow$} & \multirow{2}{*}{Metal Ion Binding $\uparrow$}& \multicolumn{2}{c}{DeepLoc $\uparrow$} \\
        \cmidrule(lr){7-8}
        Model & Loss & Modules &  &  &  & Subcellular & Binary  \\
        \cmidrule(lr){1-8}
        \multirow{4}{*}{LSD-TN-S} & \multirow{2}{*}{Univariate}  & Encoder   & 0.560 & 58.6 & 64.6 & 53.0 & 76.6 \\  
                                  &                              & Diffusion & 0.562 & 62.6 & 62.8 & 48.2 & 75.3 \\
                                  \cmidrule(lr){2-8}
                                  & \multirow{2}{*}{Multivariate}& Encoder   & 0.548 & 60.6 & 63.6 & 51.6 & 76.5 \\
                                  &                              & Diffusion & 0.528 & 53.2 & 61.1 & 44.9 & 75.3 \\
        \bottomrule
    \end{tabular}%
    }
    \caption{Normalization loss ablation:
    we compare the 
    univariate parametric form of the Kullback–Leibler divergence $\tfrac{1}{2d}\sum_i(\mu_i^2 + \sigma_i^2 - \log \sigma_i^2 -1 )$ to its multivariate counterpart $\tfrac{1}{2d}(\mu^\top \mu + \Sigma - \log\det \Sigma - d)$, with $\Sigma$ the covariance matrix. 
    }
    \label{table:eval_abl_TN}
\end{table}

\section{Context learning analysis}\label{app:attention_map}

\begin{figure}[H]
    \centering
    \includegraphics[width=1\linewidth, trim={8cm 4cm 8cm 5cm},clip]{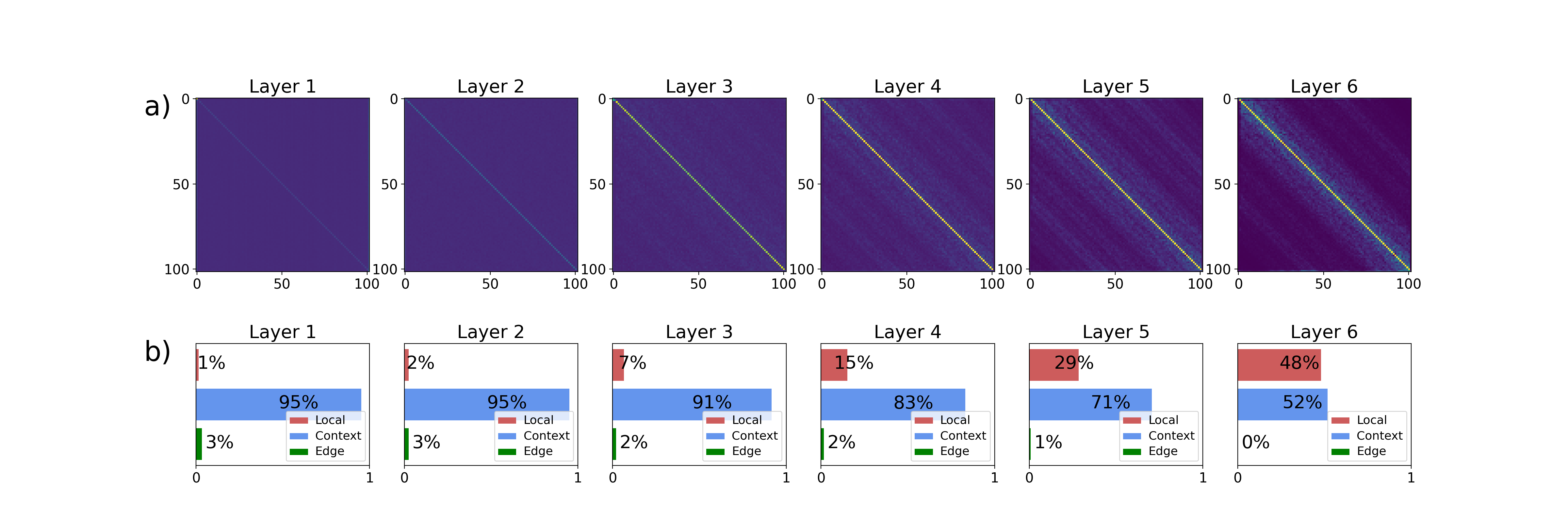}
    \caption{Attention Map Analysis for LSD-NM-S diffusion model. \textbf{a)} Average attention logits per layer, aggregated over all heads and 128 protein sequences, each consisting of 100 amino acids. \textbf{b)} Distribution of attention scores across different types: \textit{Context} attention, \textit{Local} attention, and edge-token attention.}
    \label{fig:attention_NM}
\end{figure}

\begin{figure}[H]
    \centering
    \includegraphics[width=1\linewidth, trim={8cm 4cm 8cm 5cm},clip]{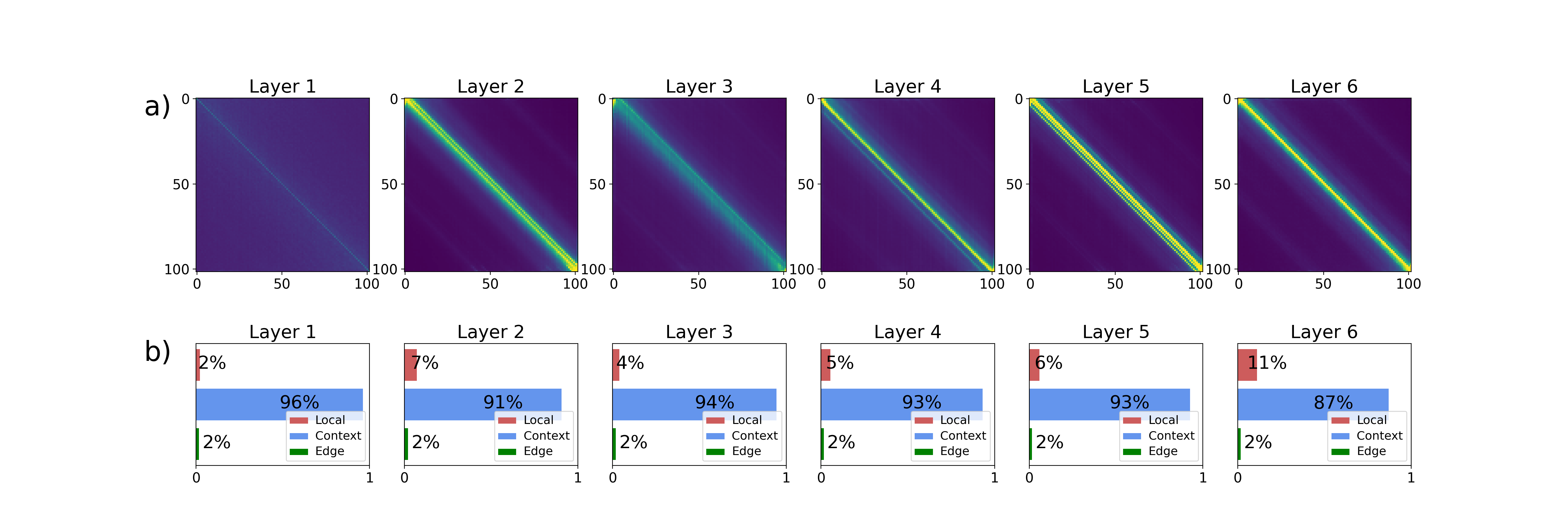}
    \caption{Attention Map Analysis for MLM-S model. \textbf{a)} Average attention logits per layer, aggregated over all heads and 128 protein sequences, each consisting of 100 amino acids. \textbf{b)} Distribution of attention scores across different types: \textit{Context} attention, \textit{Local} attention, and edge-token attention.}
    \label{fig:attention_MLM}
\end{figure}

We can better understand which elements influence a given token’s representation and how contextual information is integrated by analyzing the attention map of the transformer model and studying its distributions across the layers

In figures \ref{fig:attention_NM} and \ref{fig:attention_MLM}, we define \textit{Context} as the sum of the attention logits that connect each position in the sequence to all other different positions. \textit{Local} refers to the attention logits located along the diagonal of the attention weight matrix, representing how much a position attends to itself. Lastly, \textit{Edge} corresponds to the attention logits assigned to the EOS and BOS tokens.

For the LSD-NM-S diffusion model, the early layers predominantly focus on contextual information, with \textit{Context} attention reaching $95\%$ and \textit{Local} attention remaining as low as $1\%$. This suggests that initial layers are primarily responsible for embedding global context into each amino acid position. As the layer index increases, attention shifts to be more \textit{Local} focused, indicating that final layers refine token embeddings based on the already incorporated contextual information. The model exhibits minimal focus on edge tokens, with attention weight not exceeding $3\%$ and dropping to $0\%$ in the final layer.

In contrast, the MLM-S model maintains a strong reliance on context across all layers, consistently prioritizing \textit{Context} attention. A key difference is that attention in the MLM-S model is more short-range, with logits concentrated around nearby positions along the diagonal, whereas LSD-NM-S diffusion model distributes attention more broadly. This distinction highlights the differing information integration strategies between the two models, where LSD-NM-S diffusion model gradually transitions from global to local representation, while MLM-S persistently relies on short-range context.

\end{document}